\documentclass[runningheads]{llncs}

\usepackage[T1]{fontenc}
\usepackage{graphicx}
\usepackage{booktabs}   %
\usepackage{tabularx}   %
\usepackage{multirow}   %
\usepackage{xspace}
\usepackage[
backend=biber,
style=numeric,
sorting=ynt,
maxcitenames=50,
maxnames=50
]{biblatex}
\usepackage[inline]{enumitem}
\usepackage{tablefootnote}
\usepackage{orcidlink}
\usepackage{cleveref}

\newcommand{\MIDVHOLO}{MIDV-Holo\xspace}
\newcommand{\MIDVDYNATTACK}{MIDV-DynAttack\xspace}

\newcommand{\PUBLICURL}{\url{https://github.com/EPITAResearchLab/pouliquen.25.icdar}}

\begin{document}
\title{Verification of Dynamic Holographic Behavior\\in Identity Documents}%
\author{%
Glen Pouliquen\inst{1,2}\orcidlink{0009-0002-5231-2228} \and
Joseph Chazalon\inst{2}\orcidlink{0000-0002-3757-074X} \and\\
Guillaume Chiron\inst{1}\orcidlink{0009-0004-3665-4900} \and
Thierry G{\'e}raud\inst{2}\orcidlink{0000-0002-0380-7948} \and
Ahmad Montaser Awal\inst{1}\orcidlink{0000-0002-0479-6312}%
}
\authorrunning{Pouliquen et al.}
\institute{%
IDnow Research Center, Cesson-Sévigné, France \\
\email{name.surname@idnow.io}%
\and
EPITA Research Lab. (LRE), Le Kremlin-Bicêtre, France \\
\email{name.surname@epita.fr}
}
\maketitle
\begin{abstract} %
This paper addresses the remote verification of the authenticity of Optically Variable Devices (commonly known as holograms) on identity documents. Typically placed over the cardholder's photo, these devices provide strong and easily verifiable security for human inspection but pose challenges for automated verification.
Existing approaches easily cover static frauds (e.g. paper photocopy) and can be evaluated for such, but their capacity to detect real, dynamic fraud cases (e.g. handcrafted hologram) has not been evaluated to date because of the lack of public datasets. Furthermore, they are usually trained to detect known attack types, and few of them can generalize to new, unseen attacks.
This work features three contributions to address these limitations:
1) a new public dataset, MIDV-DynAttack, which extends the existing MIDV-Holo dataset with realistic, static and dynamic attacks against identity document specimens, tripling the number of attack samples compared to the original dataset,
2) a novel verification method which can assess the authenticity of a specific hologram thanks to the analysis of its dynamic behavior and appearance, can be trained without dynamic attack samples, and exhibits new state-of-the-art performance,
3) a benchmark of existing approaches which follows a clear evaluation protocol and emphasizes the inability of other approaches to deal with dynamic attacks, as well as new challenging attacks to deal with.
Code and dataset are publicly available at \PUBLICURL.

\end{abstract}

\begin{figure}[!tb]
\resizebox{\textwidth}{!}{
\includegraphics[]{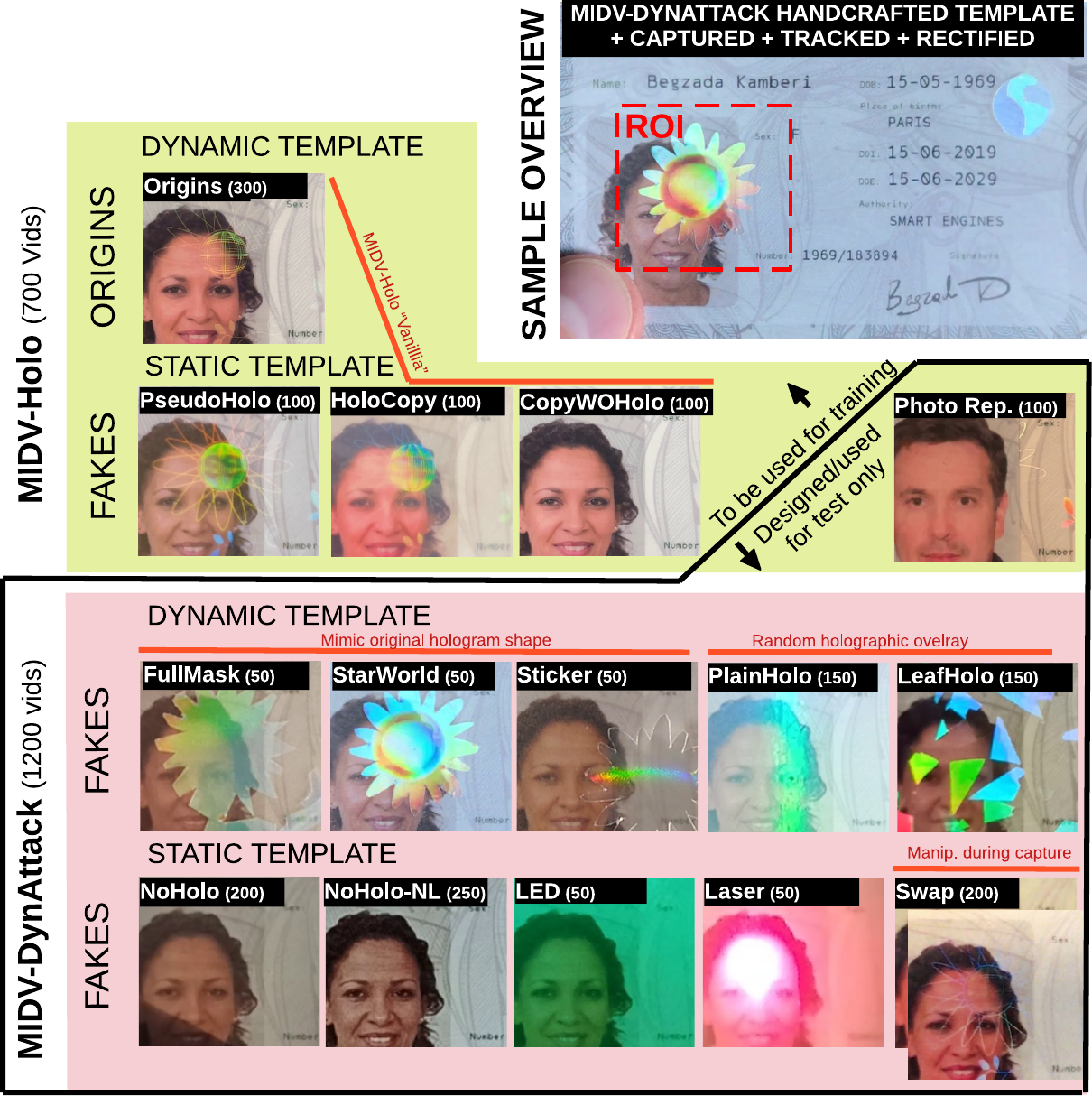}}
\caption{Our proposed dataset MIDV-DynAttack extends the original MIDV-Holo dataset with 1200 new attack videos. MIDV-DynAttack is designed for testing purposes over unseen attacks only, and not for model training or calibration.}
\label{fig:midv-holo-demo}
\end{figure}

\section{Introduction}

Many traditional identity providers still rely on physical documents for authentication, essential for scenarios like border control or restricted area access. These documents feature various security elements, both visible and invisible, to prevent forgery. However, automated remote verification is challenging due to the limited features detectable by commodity cameras on smartphones, leading to potential creative attacks that can bypass existing systems.

To address these attacks, two main directions have been explored: \textbf{Attack Detection (AD)} and \textbf{Model Verification (MV)}. AD centers on identifying subtle indicators that suggest the absence of the physical document, known as \emph{Presentation Attack Detection}, a well-established field within biometrics \cite{ramoly_pad_2024}, addressing issues like screen captures and photocopies. Additionally, AD encompasses the detection of both digital and physical manipulations, including copy-paste forgeries and replacements. This area is thoroughly documented in the literature, featuring numerous dataset publications and benchmarking competitions \cite{tapia_first_2024}.

While necessary, AD does not verify the document's consistency with its expected model. MV, the focus of this paper, checks if a document exhibits the expected security features, detecting forgeries that appear credible but do not conform to the model.

In MV, textual content is often protected by digital signatures, making tampering attempts easily detectable. However, graphical content, such as the bearer's picture, is harder to secure digitally and is usually protected by visual features like Optically Variable Devices (OVDs) or ``holograms''. These OVDs exhibit complex behaviors based on the camera, document, and light source positions, securing the bearer's picture and linking the user's face to the document.

Existing approaches for holographic content verification are insufficient for practical applications. They either verify a subset of visual appearances or check for some holographic behavior (i.e., typical hue variations for some pixels) without verifying appearance conformity, making them vulnerable to simple hologram substitutions. Additionally, no current approach assesses transitions validity, i.e., the verification of intermediate states an OVD should exhibit.

To address these limitations, we propose the following contributions:
\begin{itemize} %
    \item A new public dataset (\Cref{sec:dataset-extension}, illustrated in \Cref{fig:midv-holo-demo}) of handcrafted frauds over identity documents, focusing on the hologram region. This dataset expands the \MIDVHOLO~\cite{koliaskina_midv-holo_2023} dataset by incorporating diverse fraud scenarios, tripling the number of fraud sequences compared to the original dataset.
    \item A new method (\Cref{sec:method}) leveraging background estimation and suppression to focus on faint holographic behavior and learn a model of OVD behavior using pseudo-labels. This method effectively detects dynamic attacks previously undetected.
    \item A benchmark of existing approaches (\Cref{sec:experiments}) with a clear evaluation protocol, highlighting the most challenging attacks.
\end{itemize}
To reproduce our results, code and dataset are available at \PUBLICURL.

\section{Related Work}
\label{sec:related-work}

\subsection{Datasets} %

This section provides an overview of the available datasets for identity document fraud detection, with a specific focus on holographic content.
As summarized in Table \ref{tab:sota_datasets}, the incremental progress from genuine identity documents datasets to advanced ones based on handcrafted holographic templates can be structured into three families of datasets.

\begin{table}[!tb]
\caption{Summary of public identity document datasets related to fraud detection. HCT: HandCrafted Templates (fully synthetic or scanned source, with potential digital edition of the fields or photo). Printed HTC: acquired by a capturing device (e.g. smartphone). \uppercase{*} Preprint.%
}
\resizebox{\textwidth}{!}{%
\label{tab:sota_datasets}
\begin{tabular}{lllllll}
\hline
\textbf{Dataset} &
\begin{tabular}[c]{@{}l@{}}\textbf{No.}\\ \textbf{ids}\end{tabular} &
\begin{tabular}[c]{@{}l@{}}\textbf{Doc}\\ \textbf{types}\end{tabular} &
\begin{tabular}[c]{@{}l@{}}\textbf{Vids}\\\textbf{(+Imgs)}\end{tabular} &
\begin{tabular}[c]{@{}l@{}}\textbf{Original}\\ \textbf{source}\\ \textbf{(Legits)}\end{tabular} & \begin{tabular}[c]{@{}l@{}}\textbf{Alteration}\\ \textbf{nature}\\\textbf{(Frauds)}\end{tabular} &
\begin{tabular}[c]{@{}l@{}}\textbf{No.}\\ \textbf{Fraud}\\ \textbf{types}\end{tabular} \\
\hline
\multicolumn{7}{c}{\textit{Original genuine datasets of identity documents}} \\
\hline
\begin{tabular}[c]{@{}l@{}}MIDV-500 \cite{arlazarov_midv-500_2019} \\and 2019 \cite{bulatov_midv-2019_2020}\end{tabular} & 50 & 50 & \begin{tabular}[c]{@{}l@{}}500\\200\end{tabular} & Printed HCT & \multicolumn{2}{c}{N/A} \\
MIDV-2020 \cite{bulatov_midv-2020_2022} & 1k & 10 & \begin{tabular}[c]{@{}l@{}}1k+4k\end{tabular} & Printed HCT & \multicolumn{2}{c}{N/A} \\
DocXPand\uppercase{*} \cite{lerouge2024docxpand25k} & 25k & 15 & 0+25k & Digital HCT & \multicolumn{2}{c}{N/A} \\
BID \cite{sibgrapi_bid} & 28k & 8 & 0+28k & \begin{tabular}[c]{@{}l@{}}Digital HCT\end{tabular} &\multicolumn{2}{c}{N/A} \\
SpotBID \cite{neves_junior_doclightdetect_spotbid_2024} & 22k & 8 & 0+22k &
\begin{tabular}[c]{@{}l@{}}Digital fusion \cite{chazalon_smartdoc_nodate,sibgrapi_bid}\end{tabular} & \multicolumn{2}{c}{N/A} \\
\hline
\multicolumn{7}{c}{\textit{Fraud dataset featuring static templates (e.g. per-image digital alterations, PAD)}} \\
\hline
FMIDV \cite{ghadi_fmidv_2023} & 1k & 10 & 28k & MIDV2020 & Digital & 1\\
SIDTD \cite{boned_synthetic_2024} & 1,2k & 10 & \begin{tabular}[c]{@{}l@{}}191+8k\end{tabular} & MIDV2020 & Digital & 2\\
DLC \cite{polevoy_document_2022} & 1k & 10 & 1,4k & MIDV2020 & Recapture & 3\\
KID34k \cite{park_kid34k_2023} & 82 & 2 & 0+34k & Printed HCT & Recapture & 2 \\
IDNet \cite{guan_idnet_2024} & 41k & 20 & 0+837k & Digital HCT & Digital & 6\\
\hline
\multicolumn{7}{c}{\textit{Fraud dataset featuring dynamic templates (e.g. mimic holographic content})} \\
\hline
MIDV-Holo \cite{koliaskina_midv-holo_2023} & 100 & 20 & 700 & \begin{tabular}[c]{@{}l@{}}Printed HCT \\(+holo layer)\\\end{tabular} & \begin{tabular}[c]{@{}l@{}}Cap. altered\\template\end{tabular} & 4 \\
\begin{tabular}[c]{@{}l@{}}OUR MIDV-\\DynAttack\end{tabular} & 100 & 20 & 1200 & MIDV-Holo & \begin{tabular}[c]{@{}l@{}}Cap. altered tem-\\plates + manip.\end{tabular}&10\\
\hline
\end{tabular}}
\end{table}

\subsubsection{Original genuine datasets.}
Mobile Identity Document Video dataset (MIDV-500) \cite{arlazarov_midv-500_2019} was originally designed to address document analysis tasks (e.g. detection, classification, OCR). It was followed by multiple extensions made by the same team: MIDV-2019 \cite{bulatov_midv-2019_2020} extending capturing conditions; MIDV-2020 \cite{bulatov_midv-2020_2022} adding templates personalization; MIVD-Lait \cite{chernyshova_midv-lait_2021} featuring non-latin alphabets.
\\
The MIDV series is built over handcrafted templates, which are created from scans of identity document samples, anonymized and personalized. Personalized templates are then printed, laminated and captured using various devices (e.g. scanner, handheld smartphones). This process is especially valuable as it tends to reproduce realistic acquisition artifacts (e.g. glare, shadow, motion blur, perspective, occlusion).
\\
Afterward, MIDV-2020 was derived into different altered versions for the need of addressing specific fraud tasks. Depending on the nature the task, the original MIDV-2020 dataset can therefore be considered as genuine (e.g. when compared to digitally altered samples) or fraudulent (e.g. when compared to advanced handcrafted specimens featuring holographic content as in~\cite{pouliquen_weakly_2024}).

\subsubsection{Fraud datasets featuring static templates.}
\emph{Static templates frauds} are digital frauds that are detectable on individual frames: FMIDV~\cite{ghadi_fmidv_2023} features only crop\&move of background patches; SIDTD~\cite{boned_synthetic_2024} additionally supports inpaint\&rewrite (to alter text and signatures); IDNet~\cite{guan_idnet_2024} additionally supports text replacement, face morphing and portrait replacement.
Then, there are frauds (which can be categorized as non-compliant data) about the physical support or capturing conditions that can also be detectable on individual frames: DLC~\cite{polevoy_document_2022} and KID34k~\cite{park_kid34k_2023} datasets incorporate captures of 1) reprinted templates eventually altered (e.g. grayscale) and 2) displayed digital version of the template on a screen.
These datasets are mostly made of sequences, for which the temporal dimension is not required to address the fraud they implement. At best, the sequential aspect can benefit the final decision (Legit \emph{vs} Fraud) through some sort of aggregation of the per-image indicators, such as majority voting.

\subsubsection{Fraud datasets featuring dynamic templates.}
Finally, to address alterations of the hologram regions, there is, until now, only one dataset publicly available, MIDV-Holo~\cite{koliaskina_midv-holo_2023}, for which sequences are required to assess its legitimacy. Indeed, MIDV-Holo features advanced handcrafted dynamic templates, including a holographic layer integrated into the physical support. Less advanced static template copies (e.g., photo overlay, paper photocopy) have also been captured and labeled as frauds. This allows tasks to distinguish template nature between static and dynamic. In this context, similarly to the approach described in this paper, the original MIDV series datasets (e.g., MIDV-2020) can no longer be labeled as Legit (as in the previous task) but as Fake.

\subsection{Methods}

This section provides a brief overview of existing methods for hologram verification.
These approaches can be grouped according to whether they focus on the appearance of some OVD, its behavior, or both.

\subsubsection{Approaches looking for holographic behavior.}
To assess whether a document exhibits \emph{some} holographic behavior, \textit{Kada et~al.}~\cite{kada_hologram_2022} proposed a semantic segmentation approach, where documents captured in video were registered, and a pixel-level classification was applied to detect holograms based on color statistics.
This approach can detect whether a region contains a hologram, but it lacks any appearance classification to determine whether the expected OVD is present.
\\
\textit{Koliaskina et~al.}~\cite{koliaskina_midv-holo_2023} improved this approach by combining semantic segmentation with a global decision stage using a variance threshold, while introducing MIDV-Holo, a dataset containing real and fake holograms for evaluation.
Despite better parameterization, this approach is still not capable of discriminating between different holographic behaviors, as shown in \Cref{sec:experiments}.

\subsubsection{Approaches checking appearance conformity.}
A first approach attempting to model an OVD using limited training data is the work of \textit{Ay}~\cite{ay_open-set_2022}, which leveraged adversarial training to generate out-of-distribution samples.
This training was feasible due to the precise cropping of OVDs and their opaque nature, preventing any background from being visible through transparency and disturbing the training process.
Furthermore, as verification is performed using a binary classifier trained to discriminate between in- and out-of-distribution samples, this approach does not assess the variety of appearances a genuine OVD should exhibit and only classifies isolated frames in practice.
\\
This limitation of appearance coverage is addressed by \textit{Chapel et~al.}~\cite{chapel_authentication_2023} 
through a classifier based on Local Binary Patterns (LBP) trained on manually annotated frames, with labels indicating each particular visual appearance of some OVD.
While attractive, this approach either requires extensive annotation effort or is limited to a very restricted set of visual appearances.

\subsubsection{Approaches checking both appearance and behavior.}
The challenge of learning an accurate model of some OVD was addressed by \textit{Soukup et~al.}~\cite{soukup_mobile_2017} using dedicated acquisition hardware consisting of a smartphone connected to an LED ring, which could capture the appearance of banknote holograms under various illumination angles.
For each OVD, a CNN model is trained and used to authenticate a stack of images obtained using this dedicated hardware.
While this approach may be applicable in controlled banking environments, the use of dedicated tools is prohibitive for mass use.
\\
To check both the individual visual appearances and the variety of appearances of some OVDs using commodity smartphones, \textit{Pouliquen et~al.}~\cite{pouliquen_weakly_2024} proposed a two-stage approach that first creates an embedding of each frame, then measures the mean cosine distance of the embeddings over a video sequence.
Authentic documents produce high cosine distances due to the dynamic optical variations of genuine holograms, while fraudulent documents yield lower distances as their static reproductions fail to exhibit these variations.
The critical embedding stage is trained using a weakly supervised process, which helps the frame classification network focus on the holographic content of the image rather than the bearer's picture (which is visible because of hologram transparency).
Despite encouraging performance reported on MIDV-Holo, the appearance model learned by this approach is still weak (as shown in \Cref{sec:experiments}) as it responds to the presence of any hologram rather than the genuine OVD.

\section{MIDV-HOLO Extension with Dynamic Attacks}
\label{sec:dataset-extension}

We present \textbf{MIDV-DynAttack}, a new public dataset that significantly extends the existing MIDV-HOLO dataset with 1,200 additional video sequences.
MIDV-DynAttack is licensed under the same Creative Commons Attribution-ShareAlike conditions as the original MIDV-Holo dataset.
It is designed to challenge automatic OVD verification systems with attacks known to be problematic.

All sequences in MIDV-DynAttack were captured using one of the following smartphone models: Apple iPhone 7, Xiaomi Redmi Note 8 Pro, or Motorola G7.
Except for swap attacks, which require both hands, devices were naturally handheld to mimic realistic capturing conditions in various environments (e.g., indoor, outdoor).

Templates were handcrafted using commonly available materials and tools, i.e., MIDV-Holo templates printed, possibly laminated, or supplemented with holographic elements or layers.
We aimed to simulate realistic, easily reproducible threat scenarios that could be encountered in industrial workflows.
We did not include any deepfake samples in MIDV-DynAttack, as we consider this type of attack out of the scope of this work, which focuses on assessing how closely security features of a particular document match a known model, rather than detecting specific attacks (a complementary but different research direction).
Indeed, deepfakes require either video stream injection or screen capture, which are covered by other datasets and works (e.g., \cite{polevoy_document_2022, park_kid34k_2023}).

It is important to note that MIDV-DynAttack is meant to complement the original MIDV-Holo dataset and be used along with it.
This is why we do not provide any training or validation splits: \emph{all MIDV-DynAttack is intended to be used as a test set}, with every attack \emph{not seen during training}.
Care must be taken to avoid reusing the same identify in more than one of the training, validation or test split, and our public code cares about proper grouping.
Thanks to this protocol, MIDV-DynAttack can be effectively used as a generalization test under unknown attacks.

The MIDV-DynAttack dataset is organized into two main categories: attacks using \emph{static templates}, and the ones using \emph{dynamic templates}.
An overview of the new attacks and how they complement the original MIDV-Holo dataset is visible in \Cref{fig:midv-holo-demo}.

\subsection{Static template attacks (750 videos)}
The following categories represent traditional forgery techniques using simple printed static templates without any holographic elements. The fraudster relies on lighting effects (e.g., glare, shadows, artificial light sources producing color variations) to create dynamic content in the final video sequence, tricking the system into classifying it as holographic content. These attacks specifically target approaches that estimate variations in hue or saturation in pixel values without verifying the proper visual appearance (shape) of the OVD, or detecting and filtering light spots.

\begin{itemize}   
    \item \textbf{Natural Light (NL) - Paper (150 ID cards / 100 passports):} Template without lamination, captures were done under various lighting conditions. Very similar to the ``copy without holo'' attack from the original MIDV-Holo dataset, but with more aggressive use of light as an attack proxy.
    
    \item \textbf{Natural Light (NL) - Plastic (200 ID cards):} Template with lamination, half of the captures were done with minimal reflections (100 ID cards), and half were done with deliberate reflections from multiple light sources (100 ID cards).
        
    \item \textbf{LED-light - Plastic (50 ID cards):} Template with lamination, samples illuminated with multi-color LED-light rings to create different reflective colors.
    
    \item \textbf{Laser light - Plastic (50 ID cards):} Template with lamination, red laser randomly moving over the document.
\end{itemize}

We also added a special category that requires more than one template, each sampling a different aspect of the targeted hologram to fake. The fraudster manipulates them in front of the camera so it appears to reproduce the original dynamic holographic content. Such an attack can exhibit several valid visual appearances and aims to defeat approaches that check for multiple valid visual appearances without verifying that actual holographic behavior is present, nor actively prevent improper manipulation during the capture.
\begin{itemize}  
    \item \textbf{Swap - Paper (100 ID cards / 100 passports):} Templates without lamination. There are two different modes: swapping between two printed documents (50 ID / 100 passports) and swapping between three printed documents (50 ID).
\end{itemize}

\subsection{Dynamic template attacks (450 videos)}

The following categories represent advanced forgeries involving handcrafted dynamic templates. In this category, all templates were color printed, modified (e.g., by adding additional holographic layers or elements), and recaptured.
These attacks aim to defeat approaches that effectively look for holographic behavior, may check for a variety of visual appearances, but only loosely control the conformity of each visual appearance of the OVD.
We separated attacks with generic holographic behavior from those trying to reproduce the actual OVD shape.

\begin{itemize}   
\item Random holographic overlay (300 videos):
\begin{itemize}
    \item \textbf{Plain holo (100 ID cards / 50 passports):} Homogeneous holographic material displaying continuous diffraction patterns.
    \item \textbf{Leaf holo (100 ID cards / 50 passports):} Material exhibiting distinctive polygonal light patterns, shaped as leaves.
\end{itemize}

\item Mimicking the original holograms (150 videos):
Targeted forgeries attempting to replicate specific hologram shape and features:
\begin{itemize}
    \item \textbf{Double sticker (50 ID cards):} Combination of a flower-shaped element cut from laser holographic film and a ``genuine'' sticker (not overlapping the photo ROI).
    \item \textbf{Complete mask (50 ID cards):} Rough mask of the hologram applied over the entire printed document.
    \item \textbf{Star and world: (50 ID cards):} Combination of a main flower shape and a world-like hologram pattern.
\end{itemize}
\end{itemize}

A last point to mention is the lack of new samples that should be accepted as genuine --- we only produced new attack samples. This is due to the fact we did not have access to the original holographic layer of MIDV-Holo.
Producing a dataset with such content would be a great extension to our work.

\section{Proposed Method for OVD Verification}
\label{sec:method}
\begin{figure}[!tb]
\resizebox{\textwidth}{!}{
\includegraphics[]{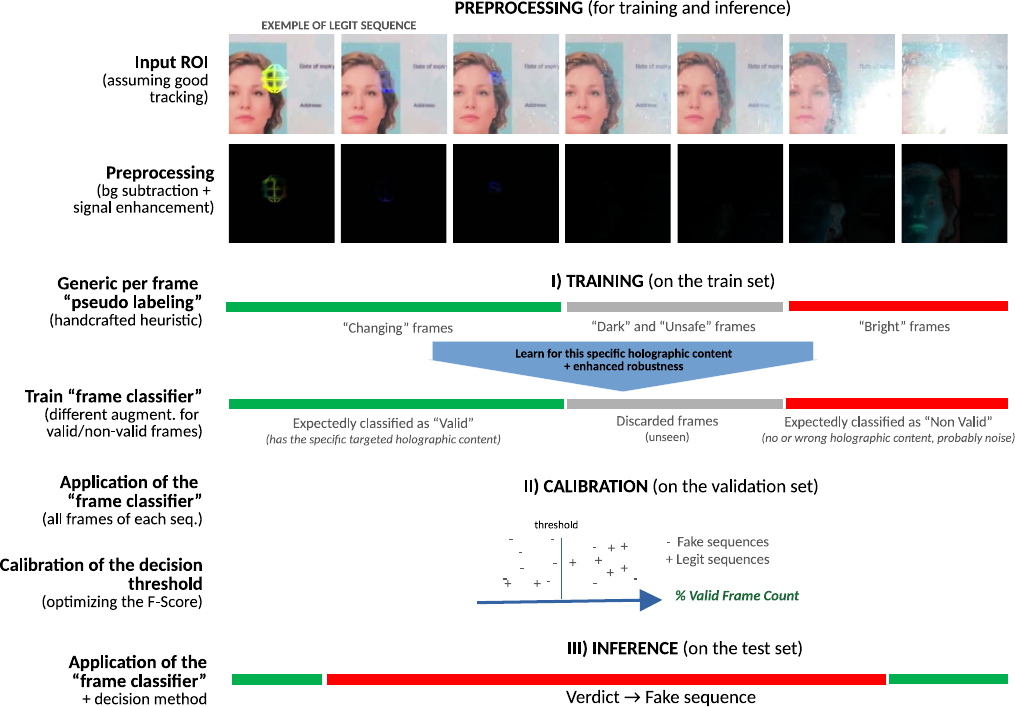}}
\caption{Overview of the proposed method. Its preprocessing step is common to both training and inference phases. The frame classifier is trained over pseudo-labels generated from a generic holographic content detector. The final decision is derived from individual frame classifications.}
\label{fig:proposed-method}
\end{figure}

The proposed method processes a video sequence of an OVD (i.e., rectified and pre-cropped to the Region Of Interest (ROI)) and produces a binary verdict (i.e., hologram Legit / Non-Legit). 
The method consists of the following key steps, as illustrated in Figure \ref{fig:proposed-method}.
First, a preprocessing step is applied to enhance the holographic signal for both the training and inference phases, primarily through background subtraction.
Next, a detection model is applied to each video frame to identify instances of genuine holographic behavior.
This detection model is trained using pseudo-labels, where frames with visible holographic behavior from Legit samples are used as positives.
Finally, the model's output undergoes a decision stage, where a threshold on the ratio of valid frames is applied to determine the sequence verdict.
This decision stage requires calibration to adapt the model's output to the specific application needs, such as varying tolerance to false rejections.

\subsection{Preprocessing to Enhance the Holographic Signal}
This preprocessing step is fully deterministic and does not require any specific parameters.
Given a previously rectified and pre-cropped OVD ROI for the input sequence, we perform two operations.
First, \textbf{background subtraction} is applied to remove the document template, including the bearer's photo, which is specific to each sequence, leaving only the variable elements.
The background image is estimated over the entire sequence using a simple per-pixel, per-channel median, and subtracted from each frame using absolute difference.
The quality of the rectification process (not described in this paper), which precisely aligns every frame of the sequence, is critical here.

Second, an HSV-based \textbf{normalizing filter} $N$ is calculated for each pixel $p$ of each frame $I_t$ as follows, with $V$ and $S$ representing the Saturation and Value channels:
\begin{equation}
N_{t}(p) = \frac{I_{t}^{(S)}(p) \cdot I_{t}^{(V)}(p)}{\max_{t',p'}\left(I_{t'}^{(S)}(p') \cdot I_{t'}^{(V)}(p')\right)},
\end{equation}
where $t'$ and $p'$ represent every timestamp and every pixel of the sequence.
This normalizing filter $N$ is applied to each original frame $I$ of the sequence (after background suppression) to produce the final signal $F$ to be used in later stages:
\begin{equation}
F_{t}^{(c)}(p) = I_{t}^{(c)}(p) \cdot N_{t}(p), \quad c \in \{R, G, B\}.
\end{equation}
This multiplication acts as a selective filter that attenuates most regions of the image while preserving only those with bright, highly saturated colors.
Areas with low Value (dark regions) are further diminished because their brightness component approaches zero.
Similarly, areas with low Saturation (regions like white, gray, or black) are reduced because their saturation component approaches zero.
Only regions exhibiting high brightness and high color saturation --- precisely the optical signature of genuine holograms --- remain prominent in the resulting image.

\subsection{Frame-level Hologram Detector}
The second step involves a trained classifier designed to distinguish authentic holographic behavior from noise and attacks.
This classifier processes a single pre-processed frame and predicts whether it is Valid or Non-Valid based on the learned holographic model.
It is trained in a supervised manner, relying on:
\begin{enumerate*}
    \item a pseudo-labeling step to generate training targets indicating whether a given frame exhibits holographic behavior;
    \item image augmentations to help the classifier capture legitimate variations of OVD and reject various forms of noise.
\end{enumerate*}

The automated \textbf{pseudo-labeling} step addresses the challenge of obtaining detailed frame-level annotations to generate training data, and labels individual frames as either \emph{Valid} or \emph{Non-Valid} according to the following process:
\begin{enumerate*}
    \item For both Legit and Non-legit sequences, the luminance ($L_t(p) \in [0, 255]$) of each preprocessed image $F_t$ is computed and thresholded to measure the amount of change $C_t = \frac{|L_t > T|}{|L_t|}$ in each frame, with $T = 5$.
    Frames are temporarily marked as \emph{``Bright''} if $C_t > \frac{1}{3}$, \emph{``Unsafe''} if $F_{t-1}$ or $F_{t+1}$ is \emph{Bright}, \emph{``Dark''} if $C_t < 1e^{-4}$ (indicating minimal change), or \emph{``Changing''} in other cases.
    \emph{Unsafe} and \emph{Dark} frames are discarded to avoid biasing the training.
    \item For \emph{Legit sequences}, \emph{Bright} frames are labeled as \emph{Non-Valid} and \emph{Changing} frames are labeled as \emph{Valid}.
    \item For \emph{Non-legit sequences} (using static template attacks from the original \MIDVHOLO dataset), both \emph{Bright} and \emph{Changing} frames are labeled as \emph{Non-Valid}.
\end{enumerate*}

Based on this supervised training set, \textbf{image augmentations} are used to teach the classifier to capture variations of valid holographic appearances while rejecting out-of-distribution samples like noise and imitation attacks.
Augmentations are applied to the pre-processed images (i.e., with background removed and after enhancement) and differ depending on whether a frame is labeled as \emph{Valid} or \emph{Non-Valid}.
For \emph{Valid} frames, we apply slight augmentations, while for \emph{Non-Valid} frames, we apply stronger augmentations as we do not fear destroying any valid signal.
Details about augmentation parameters are available in supplementary material.

\subsection{Final Sequence Classification}
The final sequence verdict (Legit / Non-Legit) is determined by applying a threshold to the proportion of frames classified as Valid within the sequence.
This threshold is calibrated on a validation set to maximize the F-score.

\section{Experiments, Results and Discussion}
\label{sec:experiments}

\subsection{Datasets and Protocol}
Our evaluation protocol closely follows the one proposed by \textit{Pouliquen et al.}~\cite{pouliquen_weakly_2024}, with the addition of our new \MIDVDYNATTACK dataset, used exclusively for testing and including attacks not seen during training.
The original \MIDVHOLO dataset is organized as follows: for each document type (identity card or passport), multiple templates are available, and for each template, 5 synthetic identities (names and face pictures) are provided.
All these documents use the same OVD layer.
To avoid any training bias on identities, we use the grouped and stratified K-fold strategy from \textit{Pouliquen et al.}~\cite{pouliquen_weakly_2024},
which averages the results over multiple runs while ensuring that the same identity is not present in more than one of the training, validation, or test splits simultaneously.

Each video is processed with a private rectification method, using original videos sampled at 15 FPS to avoid device-specific issues.
For the main experimental results, we present results at 5 FPS (one frame out of three), while additional results at the original 15 FPS can be found in the supplementary material.
This rectification improves the quality of frame alignment compared to the original \MIDVHOLO dataset and applies the same process to the new \MIDVDYNATTACK dataset, as the original tracking method was not disclosed.
We provide tracking information in the dataset for reproducibility.
Following the localization process, each document was rectified using a rigid perspective transform, and the resulting image was cropped to isolate the region of interest (ROI) containing the Optically Variable Device.
The overall structure of the rectification process is as follows, applied to both \MIDVDYNATTACK and \MIDVHOLO videos:
\begin{enumerate*}
    \item \textit{Document Localization}: Each document template is matched and localized in video frames using a detector based on local descriptors.
    \item \textit{Temporal Coherence Enhancement}: For frames where document detection was unsuccessful or unreliable, the mean between a retroactive and forward tracking approach was performed using the valid detected quads of the document.
    \item \textit{Trajectory Completion}: Any remaining untracked frames were addressed through linear interpolation between successfully localized frames.
\end{enumerate*}

All methods are trained and calibrated using the training and validation splits of the original \MIDVHOLO ``Vanilla'' dataset (which excludes the Photo Replacement attack) and tested on the rest of the dataset, ensuring identities are properly grouped.
While we report the same metrics as \textit{Pouliquen et al.}~\cite{pouliquen_weakly_2024}, which include recall, and F-score (considering a detected attack as a true positive),
we also report Area Under the ROC Curves (AUC) computed on the test set to better assess the security guarantees each method can provide.

\subsection{Methods Compared}
The following methods are tested:
\begin{enumerate}
    \item \textbf{\MIDVHOLO} method \cite{koliaskina_midv-holo_2023}, consisting of an analysis of pixel-wise chromaticity statistics accumulated in a video stream. We use the public implementation from~\cite{pouliquen_weakly_2024}.
    
    \item \textbf{Weakly Supervised Learning} (WSL) \cite{pouliquen_weakly_2024} using contrastive learning and the label at the video level to optimize a feature extraction and frame-level classification pipeline, previously producing state-of-the-art results.

    \item \textbf{Direct Classifier} refereed as Naive Classifier in \cite{pouliquen_weakly_2024}, a simple binary frame classifier leveraging any categorizable signal (e.g. print artifacts).
    It can rely on any image signal (including unexpected biases) from the original \MIDVHOLO dataset to discriminate genuine and attack sequences based on isolated frame classification.

    \item \textbf{HoloVerif} (our proposed method), involving background subtraction pre-processing and pseudo-labeling to train a frame classifier in a supervised manner, focusing on holographic content (as detailed in \Cref{sec:method}). It is specifically designed to improve support for dynamic template attacks.
\end{enumerate}

To measure the impact of each stage of our proposed approach, we gradually integrate them in an ablation study.
We first tried to assess whether some simple binary frame classifier (Direct Classifier method), as in \cite{pouliquen_weakly_2024}, could be applied on preprocessed frames (called ``With bg. subtraction'') to accurately and directly predict the sequence class, using only global (potentially biased) Legit/Non-legit targets for training.
We also measured the classification power of the hardcoded preprocessing (called ``Pseudo labels only''), using the same decision function at the sequence level (threshold over the ratio of Valid frames).
Furthermore, we benchmarked a version of our proposed method HoloVerif without augmentations (called ``No augmentations'') .
Training details for each method are provided as supplementary material.

\subsection{Results and Discussion}

\begin{table}[!tb]
    \centering
    \caption{Results on MIDV-Holo and MIDV-DynAttack datasets reported as mean ± standard deviation over 5-fold cross-validation. For methods including randomness, results are averaged across 6 different random seeds. The frames were sampled at 5 FPS. F-score and AUC metrics are reported on subsets containing both Legit and Fraud sequences, while Recall is reported only on Fraud-only subsets. The numbers in bold highlight the discriminative strength of the methods.}
    \label{tab:results}
    \resizebox{\textwidth}{!}{%
    \begin{tabular}{lccccccc}
        \toprule
        & \multicolumn{3}{c}{\textbf{\MIDVHOLO} (test split)} & \multicolumn{3}{c}{\textbf{MIDV-DynAttack} (full)} & \textbf{Both} \\
        \cmidrule(lr){2-4} \cmidrule(lr){5-7} \cmidrule(lr){8-8}
        & \multicolumn{2}{c}{\textbf{Vanilla}} & \textbf{Photo Rep.} & \textbf{Static} & \textbf{Static-swap} & \textbf{Dynamic} & \textbf{Mix} \\
        \multirow{3}{*}{\begin{tabular}[c]{@{}l@{}}\textbf{Verification}\\\textbf{methods $\downarrow$}\end{tabular}}
        & \multicolumn{2}{c}{120 mixed vids} & 20 vids & 110 vids & 40 vids & 90 vids & 380 vids \\
        \cmidrule(lr){2-3} \cmidrule(lr){4-4} \cmidrule(lr){5-5} \cmidrule(lr){6-6} \cmidrule(lr){7-7} \cmidrule(lr){8-8}
        & \textbf{F-score} & \textbf{Recall} & \textbf{Recall} & \textbf{Recall} & \textbf{Recall} & \textbf{Recall} & \textbf{AUC} \\
        \midrule
        \multicolumn{8}{c}{\textit{Hologram detection (no verification), $\ddagger$ low Dynamic scores by design}} \\
        \midrule
        \MIDVHOLO \cite{koliaskina_midv-holo_2023} & 85 ± 3 & 85 ± 12 & 38 ± 21 & 72 ± 3 & 19 ± 8 & 3$^\ddagger$ ± 3 & 57 ± 1\\
        WSL \cite{pouliquen_weakly_2024} & \textbf{96 ± 2} & 97 ± 4 & \textbf{92 ± 8} & 81 ± 8 & 14 ± 8 & 2$^\ddagger$ ± 2 & 83 ± 3\\
        \midrule
        \multicolumn{8}{c}{\textit{Generic fraud detector, leveraging any categorizable signal (e.g. print artifacts, biases)}} \\
        \midrule
        Direct Classifier \cite{pouliquen_weakly_2024} & 90 ± 4 & 89 ± 6 & 70 ± 17 & 82 ± 11 & \textbf{90 ± 9} & 52 ± 20  & \textbf{92 ± 3}\\
        \midrule
        \multicolumn{8}{c}{\textit{Hologram Verification (focusing more on holographic content using background subtraction)}} \\
        \midrule
        HoloVerif (ours) & \textbf{95 ± 3} & 96 ± 5 & 59 ± 20 & \textbf{93 ± 5} & 28 ± 12 & \textbf{61 ± 13} & \textbf{93 ± 2}\\
    
        \bottomrule
        \\
        \multicolumn{8}{c}{\textbf{Ablation study}, $\dagger$ showing strong negative impact.}\\
        \midrule
        \multicolumn{8}{c}{\textit{For HoloVerif (ours)}}\\
        \midrule
        Pseudo labels only & 64$^\dagger$± 4 & 80 ± 14 & 71 ± 20 & 87 ± 8 & 60 ± 22 & 65 ± 13  & 61 ± 2\\
        No augmentations & \textbf{96 ± 3} & 96 ± 4 & 41 ± 11 & \textbf{94 ± 4} & 19 ± 9 & 29$^\dagger$± 10 &  87 ± 3\\
        \midrule
        \multicolumn{8}{c}{\textit{ For Direct Classifier \cite{pouliquen_weakly_2024}}}\\
        \midrule
        With bg. subtraction & \textbf{97 ± 2} & 99 ± 2 & 47$^\dagger$± 16 & 81 ± 9 & 17 ± 11 & 21$^\dagger$± 14 & 89 ± 3\\        
        \bottomrule
    \end{tabular}
    } %
\end{table}

\begin{figure}[!tb]
\resizebox{\textwidth}{!}{
\includegraphics[]{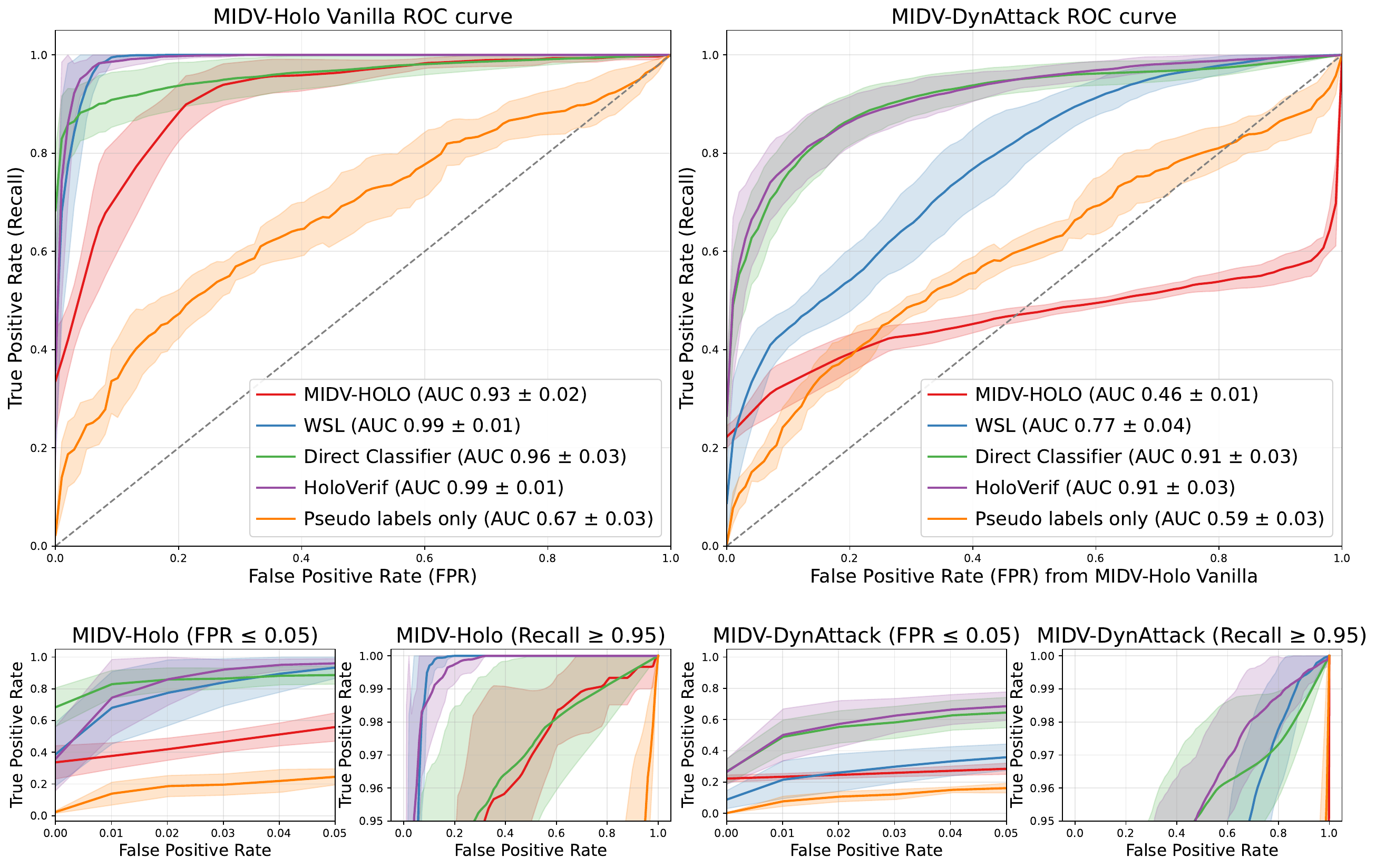}}
\caption{ROC curve of selected methods for \MIDVHOLO ``Vanilla'' (left) and \MIDVDYNATTACK (right) test sets, after calibration on a separate set. Bottom row shows zoomed regions for either low false positive rate (easier to use), or high true positive rate (safer) regimes. A \emph{True Positive} is a fraud sample properly detected. The transparent area around each curve indicates the variance ($\pm\sigma$) observed over 5 folds ($\times$ 6 seeds for methods with randomness). Curves were produced by varying the value of the final decision threshold.}
\label{fig:roccurve}
\end{figure}

Experimental results are summarized in \Cref{tab:results} for all methods and ablation studies, separating the metrics for the original \MIDVHOLO and our new \MIDVDYNATTACK datasets.
\Cref{fig:roccurve} provides complementary insights regarding ideal decisions which could be made on the test sets, to better identify similarities and the security vs ease-of-use tradeoff each method can offer.

\MIDVDYNATTACK presents a more challenging evaluation environment, as shown by the general drop in performance between the original \MIDVHOLO dataset and the proposed one.
As shown in the general difference between left and right ROC curves in \Cref{fig:roccurve}.
In particular, \MIDVHOLO method performance drops by 10 points between static fraud cases in original \MIDVHOLO and \MIDVDYNATTACK.
The +5 points improvement on the \MIDVHOLO dataset of the \MIDVHOLO method (applied to the photo region), compared to results reported in~\cite{pouliquen_weakly_2024}, are attributable to our enhanced tracking.

The Direct Classifier on original images focuses primarily on general image characteristics rather than hologram elements. Our ablation study confirms that preprocessing reduces these biases, with the Direct classifier showing decreased performance (90\% to 17\% on static swap) after background subtraction, indicating a partial destruction of these hidden image features.
Our HoloVerif approach deliberately focuses on holographic content, and reaching leading performance on \MIDVHOLO, as well as static and dynamic unseen attacks from \MIDVDYNATTACK.
Augmentations substantially improve performance on dynamic attacks, from 29\% to 61\% recall.
WSL still achieve a marginally superior performance for the static attacks of the original \MIDVHOLO dataset.

Two attacks are particularly challenging for almost all methods.
Performance on the Photo Replacement attack is low for all methods except WSL, and is due to the fact some part of the hologram remain visible outside the replaced face picture.
Further study of the WSL method would be required to understand this effect.
This vulnerability suggests that precisely focusing on critical elements like face picture is essential.
The Document Swap attack is also very challenging: the \MIDVHOLO and WSL method may be tricked by the multiple saturated patterns passing their variety control, and HoloVerif is defeated because of its background estimation which estimate the background frame as one of the templates shown; the others exhibiting a difference image with acceptable holographic appearance.
This second vulnerability may be combated in several ways: including Attack Detection targeting swapping is a natural option, but it could also be addressed by including broader signals (as does the Direct Classifier) or, more interestingly, improve hologram modeling with a temporal aspect, i.e., learning instant color variations from the ordered sequence instead of processing it as a set of frames.

Overall, the Direct Classifier method and our proposed HoloVerif methods have superior performance regarding aggregated AUC metric (> 90), though they do not detect the same attacks.
The Direct Classifier is good at detecting static Frauds, and catches some dynamic ones as well, eventually thanks to difference in the image signal between legit and non-legit cases caught during the training phase.
This could complement our proposed method which exhibits leading performance on Dynamic attacks.

\section{Conclusion}
This paper tackles the challenge of remotely verifying the authenticity of Optically Variable Devices (OVDs) on identity documents under realistic conditions.
We introduced \MIDVDYNATTACK, a test-only dataset extending \MIDVHOLO with new static (showing a single OVD appearance) and dynamic (showing realistic but altered holographic behavior) attack scenarios. It is the first public dataset featuring attacks of this sophistication.
Experiments demonstrated that existing methods are vulnerable to carefully crafted dynamic attacks.
To address this, we proposed \emph{HoloVerif}, a method that leverages background estimation to isolate the holographic signal. It achieved state-of-the-art performance on dynamic attacks while maintaining strong results on static ones.
Despite these advances, defeating dynamic attacks remains a challenge. Improved training strategies are needed to boost generalization and robustness.
We also identified two particularly difficult physical attacks:
\textbf{Photo Replacement}, due to authentic OVD elements surrounding the image, and
\textbf{Document Swapping}, which misleads systems relying on multiple static OVD views.
Addressing these issues requires enhanced Model Verification (MD) systems that:
\begin{enumerate*}
\item focus on critical areas like the bearer's photo;
\item model OVD temporal dynamics by analyzing frame sequences in context rather than as unordered images.
\end{enumerate*}
Additionally, production systems should integrate MD with Attack Detection (AD) to counter known threats like Document Swapping.
Finally, while we assessed recall on unseen attacks, we lacked new genuine samples to evaluate generalization precision.
We encourage the community to expand our open dataset with more genuine instances and to reuse our public evaluation code.

\begin{credits}
\subsubsection{\ackname} We thank Vincent Léon for his assistance with video recording of documents for the dataset.
The ACHILLE project, partially supporting this work, was funded by the European Union’s Horizon Europe research and innovation program under Grant Agreement No 101189689.

\subsubsection{\discintname}
The authors have no competing interests.
\end{credits}

\printbibliography

@INPROCEEDINGS{ramoly_pad_2024,
  author={Ramoly, Nathan and Komaty, Alain and Hahn, Vedrana Krivokuća and Younes, Lara and Awal, Ahmad-Montaser and Marcel, Sébastien},
  booktitle={Proceedings of the IEEE International Joint Conference on Biometrics ({IJCB})}, 
  title={A Novel and Responsible Dataset for Face Presentation Attack Detection on Mobile Devices}, 
  year="2024",
  volume={},
  number={},
  pages={1-9},
  doi={10.1109/IJCB62174.2024.10744500}}

@article{polevoy_document_2022,
	title = {Document Liveness Challenge Dataset ({DLC}-2021)},
	volume = {8},
	doi = {10.3390/jimaging8070181},
	pages = {181},
	number = {7},
	journaltitle = {Journal of Imaging},
	author = {Polevoy, Dmitry V. and Sigareva, Irina V. and Ershova, Daria M. and Arlazarov, Vladimir V. and Nikolaev, Dmitry P. and Ming, Zuheng and Luqman, Muhammad Muzzamil and Burie, Jean-Christophe},
	date="2022"
}

@article{arlazarov_midv-500_2019,
	title = {{MIDV}-500: A Dataset for Identity Documents Analysis and Recognition on Mobile Devices in Video Stream},
	volume = {43},
	doi = {10.18287/2412-6179-2019-43-5-818-824},
	number = {5},
	journaltitle = {Computer Optics},
	author = {Arlazarov, Vladimir V. and Bulatov, Konstantin and Chernov, Timofey and Arlazarov, Vladimir L.},
	date = {2019}
}

@inproceedings{bulatov_midv-2019_2020,
	title = {{MIDV}-2019: Challenges of the modern mobile-based document {OCR}},
	doi = {10.1117/12.2558438},
	pages = {64},
	booktitle = {Proceedings of the twelfth International Conference on Machine Vision ({ICMV} 2019)},
	author = {Bulatov, Konstantin and Matalov, Daniil and Arlazarov, Vladimir V.},
	date = {2020}
}

@article{bulatov_midv-2020_2022,
	title = {{MIDV}-2020: A Comprehensive Benchmark Dataset for Identity Document Analysis},
	volume = {46},
	doi = {10.18287/2412-6179-CO-1006},
	number = {2},
	journaltitle = {Computer Optics},
	author = {Bulatov, Konstantin and Emelianova, Ekaterina and Tropin, Daniil and Skoryukina, Natalya and Chernyshova, Yulia and Sheshkus, Alexander and Usilin, Sergey and Ming, Zuheng and Burie, Jean-Christophe and Luqman, Muhammad Muzzamil and Arlazarov, Vladimir V.},
	date = {2022}
}

@inproceedings{guan_idnet_2024,
	title = {IDNet: A Novel Dataset for Identity Document Analysis and Fraud Detection},
	author = {Xie, Lulu and Wang, Yancheng and Guan, Hong and Nag, Soham and Goel, Rajeev and Swamy, Niranjan and Yang, Yingzhen and Xiao, Chaowei and Prisby, Jonathan and Maciejewski, Ross and Zou, Jia},
	date = {2024},
    pages = {2244-2253},
    doi = {10.1109/BigData62323.2024.10825017},
    booktitle = {Proceedings of the IEEE International Conference on Big Data ({BigData}) },
}

@incollection{chernyshova_midv-lait_2021,
	title = {{MIDV}-{LAIT}: A Challenging Dataset for Recognition of {IDs} with Perso-Arabic, Thai, and Indian Scripts},
	pages = {258--272},
    booktitle = {Proceedings of the 16th International Conference Document Analysis and Recognition.},
	author = {Chernyshova, Yulia and Emelianova, Ekaterina and Sheshkus, Alexander and Arlazarov, Vladimir},
	date = {2021},
	doi = {10.1007/978-3-030-86331-9_17},
}

@article{boned_synthetic_2024,
  title={Synthetic dataset of id and travel documents},
  author={Boned, Carlos and Talarmain, Maxime and Ghanmi, Nabil and Chiron, Guillaume and Biswas, Sanket and Awal, Ahmad Montaser and Ramos Terrades, Oriol},
  journal={Scientific Data},
  volume={11},
  number={1},
  pages={1356},
  year="2024",
  doi={10.1038/s41597-024-04160-9},
  publisher={Nature Publishing Group UK London}
}

@inproceedings{ghadi_fmidv_2023,
  author={Al-Ghadi, Musab and Ming, Zuheng and Gomez-Krämer, Petra and Burie, Jean-Christophe and Coustaty, Mickaël and Sidere, Nicolas},
  booktitle={Proceedings of the 25th International Workshop on Multimedia Signal Processing}, 
  title={Guilloche Detection for ID Authentication: A Dataset and Baselines}, 
  year={2023},
  volume={},
  number={},
  pages={1-6},
  doi={10.1109/MMSP59012.2023.10337681}
}

@misc{lerouge2024docxpand25k,
  title={DocXPand-25k: a large and diverse benchmark dataset for identity documents analysis}, 
  author={Julien Lerouge and Guillaume Betmont and Thomas Bres and Evgeny Stepankevich and Alexis Bergès},
  year={2024},
  eprint={arXiv/2408.01690},
}

@inproceedings{sibgrapi_bid,
 author = {Álysson Soares and Ricardo das Neves Junior and Byron Bezerra},
 title = {BID Dataset: a challenge dataset for document processing tasks},
 booktitle = {Proceedings of the XXXIII Conference on Graphics, Patterns and Images},
 year = {2020},
 pages = {143--146},
 doi = {10.5753/sibgrapi.est.2020.12997}
}

@inproceedings{park_kid34k_2023,
	title = {KID34K: A Dataset for Online Identity Card Fraud Detection},
	doi = {10.1145/3583780.3615122},
	pages = {5381--5385},
	booktitle = {Proceedings of the 32nd {ACM} International Conference on Information and Knowledge Management},
	author = {Park, Eun-Ju and Back, Seung-Yeon and Kim, Jeongho and Woo, Simon S.},
	year="2023",
}

@inproceedings{tapia_first_2024,
	title = {First Competition on Presentation Attack Detection on {ID} Card},
	doi = {10.1109/IJCB62174.2024.10744475},
	pages = {1--10},
	booktitle = {Proceedings of the {IEEE} International Joint Conference on Biometrics ({IJCB})},
	author = {Tapia, Juan E. and Damer, Naser and Busch, Christoph and Espin, Juan M. and Barrachina, Javier and Rocamora, Alvaro S. and et al.},
	date = "2024"
}

@inproceedings{neves_junior_doclightdetect_spotbid_2024,
	title = {DocLightDetect: A New Algorithm for Occlusion Classification in Identification Documents},
	doi = {10.1007/978-3-031-70442-0_12},
	pages = {196--210},
	booktitle = {Proceedings of the 16th {IAPR} International Workshop on Document Analysis Systems ({DAS})},
	author = {das Neves Junior, Ricardo Batista and Dantas Bezerra, Byron Leite and Zanchettin, Cleber},
	year = "2024",
}

@inproceedings{chazalon_smartdoc_nodate,
  author={Chazalon, J. and Gomez-Krämer, P. and Burie, J.-C. and Coustaty, M. and Eskenazi, S. and Luqman, M. and Nayef, N. and Rusiñol, M. and Sidère, N. and Ogier, J.-M.},
  booktitle={Proceedings of the 14th IAPR International Conference on Document Analysis and Recognition (ICDAR)}, 
  title={SmartDoc 2017 Video Capture: Mobile Document Acquisition in Video Mode}, 
  year={2017},
  volume={04},
  number={},
  pages={11-16},
  doi={10.1109/ICDAR.2017.306}
}

@inproceedings{koliaskina_midv-holo_2023,
author="Koliaskina, L. I.
and Emelianova, E. V.
and Tropin, D. V.
and Popov, V. V.
and Bulatov, K. B.
and Nikolaev, D. P.
and Arlazarov, V. V.",
title="MIDV-Holo: A Dataset for ID Document Hologram Detection in a Video Stream",
booktitle="Proceedings of the 17th International Conference on Document Analysis and Recognition (ICDAR)",
year="2023",
pages="486-503",
doi = {10.1007/978-3-031-41682-8_30},
}

@article{ay_open-set_2022,
	title = {Open-Set Learning-Based Hologram Verification System Using Generative Adversarial Networks},
	volume = {10},
	doi = {10.1109/ACCESS.2022.3155870},
	pages = {25114--25124},
	journaltitle = {{IEEE} Access},
	author = {Ay, Betul},
	date = {2022},
}

@article{soukup_mobile_2017,
	title = {Mobile hologram verification with deep learning},
	volume = {9},
	doi = {10.1186/s41074-017-0022-7},
	journaltitle = {{IPSJ} Transactions on Computer Vision and Applications},
	shortjournal = {{IPSJ} Transactions on Computer Vision and Applications},
	author = {Soukup, Daniel and Huber-Mörk, Reinhold},
	date="2017",
}

@inproceedings{kada_hologram_2022,
	title = {Hologram Detection for Identity Document Authentication},
	doi = {10.1007/978-3-031-09037-0_29},
	pages = {346--357},
	booktitle = {Proceedings of the 3rd International Conference on Pattern Recognition and Artificial Intelligence},
	author = {Kada, Oumayma and Kurtz, Camille and van Kieu, Cuong and Vincent, Nicole},
	date = {2022},
}

@inproceedings{chapel_authentication_2023,
	title = {Authentication of Holograms with Mixed Patterns by Direct {LBP} Comparison},
	doi = {10.1109/MMSP59012.2023.10337669},
	pages = {1--6},
	booktitle = {Proceedings of the 25th International Workshop on Multimedia Signal Processing},
	author = {Chapel, Marie-Neige and Al-Ghadi, Musab and Burie, Jean-Christophe},
	date = {2023},
}

@inproceedings{pouliquen_weakly_2024,
author = {Pouliquen, Glen and Chiron, Guillaume and Chazalon, Joseph and G\'{e}raud, Thierry and Awal, Ahmad Montaser},
title = {Weakly Supervised Training for Hologram Verification in Identity Documents},
year = {2024},
doi = {10.1007/978-3-031-70533-5_2},
booktitle = {Proceedings of the 18th International Conference on Document Analysis and Recognition (ICDAR)},
pages = {17–33},
numpages = {17},
}

\clearpage
\appendix
\makeatletter
\newcommand{\thetitle}{}
\let\titleold\title
\renewcommand{\title}[1]{\titleold{#1}\gdef\thetitle{#1}}
\makeatother

\makeatletter
\newcommand{\maketitlesupplementary}{%
  \clearpage
  \thispagestyle{plain}%
  \begin{center}
    {\LARGE\bfseries \thetitle\par}%
    \vspace{0.5em}%
    {\large Supplementary Material\par}%
    \vspace{1.0em}%
  \end{center}%
}
\makeatother

\maketitlesupplementary

\counterwithin{figure}{section}
\counterwithin{table}{section}

This supplementary material contains the following sections:

\begin{itemize}
    \item \textbf{\Cref{app:dataset-viz} - Dataset Visualization}:
    Illustrations of the MIDV-Holo and MIDV-DynAttack datasets showing rectified document samples.

    \item \textbf{\Cref{app:dataset-crea} - Dataset Creation}:
    Description of the capture conditions.

    \item \textbf{\Cref{app:training} - Training Methodology}: Detailed explanation of model architecture, preprocessing, and class-specific augmentation strategies.

    \item \textbf{\Cref{app:results-aggre} - Results Aggregation}: Explanation of the cross-validation protocol and the Area Under the Curve computation.
    
    \item \textbf{\Cref{app:results-15fps} - High Frame Rate Analysis}:
    Extended experimental results at 15 FPS complementing the 5 FPS results in the main paper.

    \item \textbf{\Cref{app:per-attack} - Per-Attack Performance Analysis}: Performance for each individual fraud type in MIDV-DynAttack.
    
    \item \textbf{\Cref{app:hogsgd} - Lightweight Implementation}: Evaluation of a computationally efficient HOG-SGD variant of the proposed method.
    
\end{itemize}
\clearpage
\section{Dataset Visualization}
\label{app:dataset-viz}

\begin{figure}[!h]
\resizebox{\textwidth}{!}{
\includegraphics[]{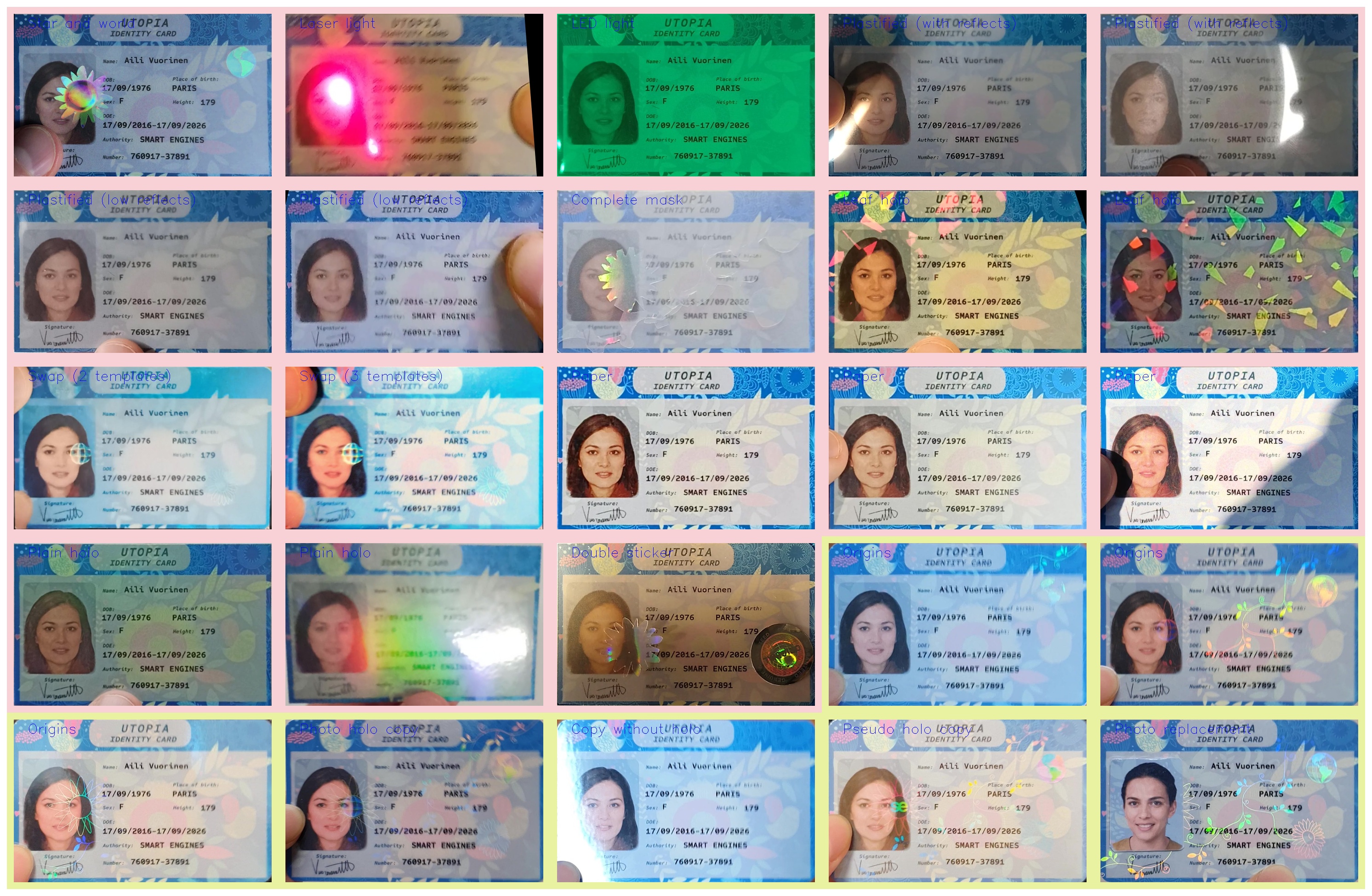}}
\caption{Overview of the first frame for all captures of the second ID document type, fifth identity. Frame samples from \MIDVHOLO are highlighted with green borders, while samples from \MIDVDYNATTACK are indicated with red borders.}
\label{fig:rectified-overview}
\end{figure}

\begin{figure}[!h]
\resizebox{\textwidth}{!}{
\includegraphics[]{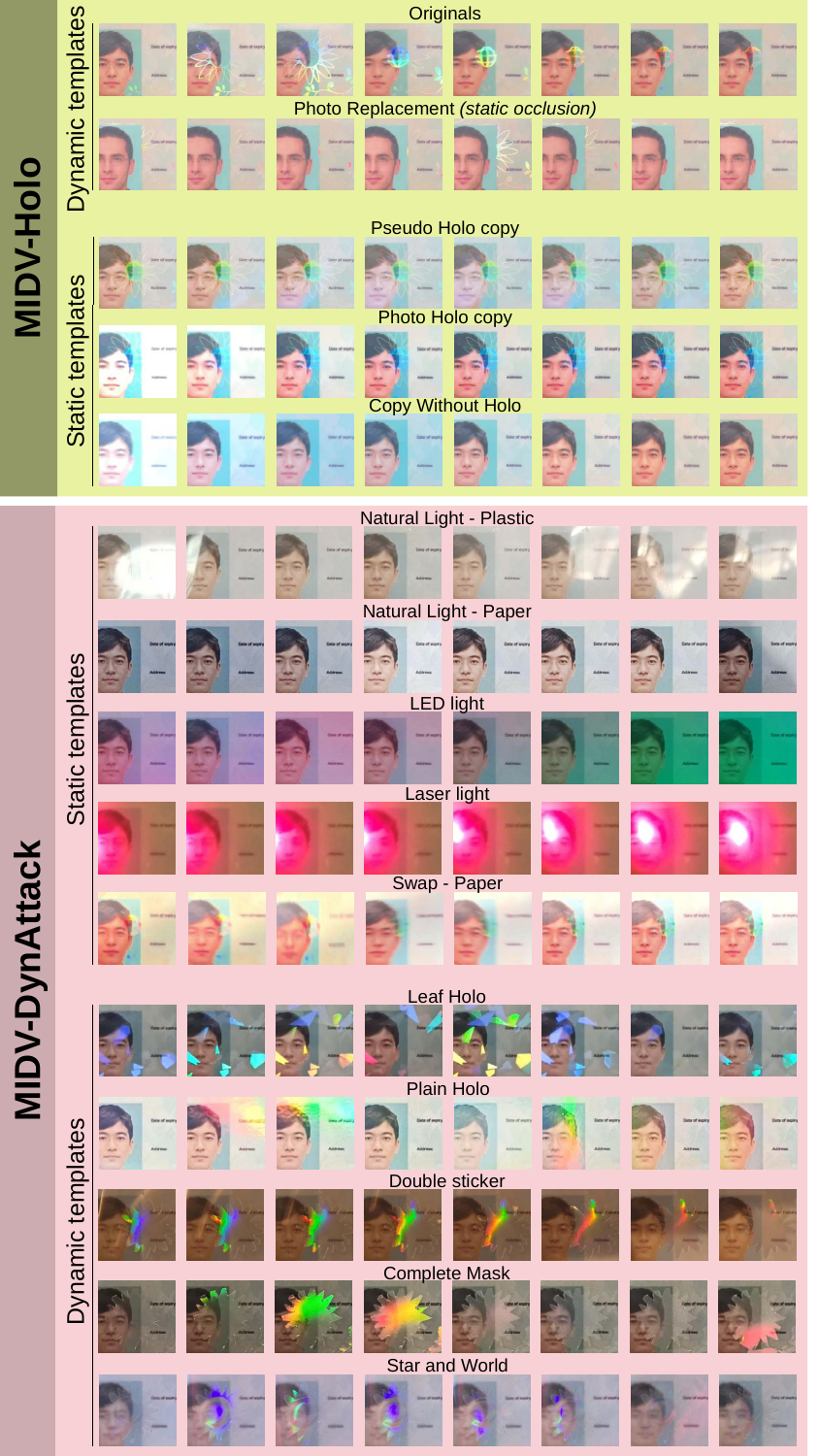}}
\caption{Overview of sequences for all fraud types of the fifth type of ID documents, first identity. Frame samples from \MIDVHOLO are highlighted with green borders, while samples from \MIDVDYNATTACK are indicated with red borders.}
\label{fig:roi-sequence-overview}
\end{figure}

The \Cref{fig:rectified-overview} shows an overview of the MIDV-Holo and MIDV-DynAttack datasets, in the form of a mosaic made of the first sample from all captured videos, each being rectified using document coordinates provided in the annotations. The background is therefore not visible in the illustration.
The \Cref{fig:roi-sequence-overview} shows an overview of sequences, displaying several frames for each video.

\clearpage

\clearpage

\section{Dataset Capture}
\label{app:dataset-crea}

The dataset was captured at 30 FPS using three smartphone models: an iPhone 7, a Xiaomi Redmi Note 8 Pro, and a Motorola G7.
To simulate real-world KYC scenarios, most videos were recorded in naturalistic conditions (except ``LED light'' and ``Laser light''). We deliberately varied capture distances and lighting to reflect uncontrolled authentication environments.

\clearpage

\section{Training Methodology}
\label{app:training}
The training strategy for the methods is as follows: input ROIs resized to 230 pixels (256 for WSL and Direct Classifier), followed by random cropping to $224 \times 224$ pixels. After applying method-specific transformations, inputs were normalized according to ImageNet statistics for original cropped images and to mean 0.1 and standard deviation 0.2 for preprocessed images. All deep learning methods used the $mobilevit_{xxs}$ architecture, optimized using AdamW (learning rate = 1e-4) with batch sizes of 128 for classifier-based approaches and 32 for WSL.

For the proposed frame-level HoloVerif method, we implemented a class-dependent augmentation pipeline.\\
For \emph{Valid} frames, we apply slight augmentations: resize to 230px, then random crop to 224px. Color adjustments include color jitter with brightness variation ±0.3, contrast variation ±0.3, saturation variation ±0.2, and hue variation ±0.1. Finally, normalize to mean=0.1 and std=0.2.
\\
For \emph{Non-Valid} frames, we apply stronger augmentations as we do not fear destroying any valid signal: resize to 230px, then random crop to 224px, with a probability of 0.2 apply a sequence of: horizontal flip (p=0.5), vertical flip (p=0.3), random rotation (up to 30°), elastic transform (p=0.3, alpha=60, sigma=6), non-linear deformation (p=0.3, gridsize=4, magnitude=0.25), and affine transformation (degrees=30, translate=±0.2, scale=0.8-1.2, shear=20°). Color adjustments include color jitter with brightness variation ±0.4, contrast variation ±0.4, saturation variation ±0.3, and hue variation ±0.4. Finally, normalize to mean=0.1 and std=0.2.

\clearpage

\section{Results Aggregation}
\label{app:results-aggre}

\subsection{Cross-validation protocol and statistical analysis}
We employed a 5-fold cross-validation strategy for all methods. The evaluation protocol differs based on whether the method incorporates stochastic components. For deterministic methods (\MIDVHOLO baseline and pseudo-labels only), we conducted 5 runs corresponding to each fold. For methods involving randomization (all deep learning approaches), we performed 30 runs in total, combining 5 folds with 6 different random seeds.

The final performance metrics are reported as mean ± standard deviation computed across all runs. For categorical analysis, we first aggregated results within each fraud category for individual runs, then computed statistical measures across multiple runs.

We also report an inconsistency in the \MIDVHOLO dataset: the document $id03\_03$ from the originals collection was not captured and was replaced by captures of $id07\_03$. Since this document appeared twice in the collection, we retained only one instance to avoid duplication, resulting in one document without the corresponding three video sequences in the originals collection.

\subsection{Area Under the Curve computation}
Our AUC computation follows a systematic two-stage approach. First, we computed ROC curves using MIDV-Holo Vanilla samples, which contain both legitimate and fraudulent documents, establishing the baseline false positive rate (FPR) and true positive rate (TPR) relationship. We then defined 100 evenly distributed thresholds and extracted the corresponding FPR values from these ROC curves.

For each fraud type in MIDV-DynAttack, we computed recall values at these predefined thresholds and calculated individual AUC scores using trapezoidal integration. The final reported AUC represents a weighted average based on the number of video sequences in each category, ensuring that the aggregated metrics reflect the relative importance of each dataset component.

\clearpage

\section{High Frame Rate Analysis}
\label{app:results-15fps}
All video frames were extracted at 15 FPS using the FFmpeg program with the command: \textit{ffmpeg -i vid.mp4 -vf fps=15 -q:v 2 img\_\%04d.jpg}. In the main manuscript, we subsampled these extracted frames to 5 FPS to ensure a fair comparison with previously published methods. The results using the full 15 FPS sequences (without subsampling) are presented in \Cref{tab:results15fps}. 
Trends are similar to the results reported in the main paper at 5 FPS.

\begin{table}[!h]
    \centering
    \caption{Results at \textbf{15fps} on MIDV-Holo and MIDV-DynAttack datasets reported as mean ± standard deviation over 5-fold cross-validation. For methods including randomness, results are averaged across 6 different random seeds. The frames were sampled at 15 FPS. F-score and AUC metrics are reported on subsets containing both Legit and Fraud sequences, while Recall is reported only on Fraud-only subsets. The numbers in bold highlight the discriminative strength of the methods.}
    \label{tab:results15fps}
    \resizebox{\textwidth}{!}{%
    \begin{tabular}{lccccccc}
        \toprule
        & \multicolumn{3}{c}{\textbf{\MIDVHOLO} (test split)} & \multicolumn{3}{c}{\textbf{MIDV-DynAttack} (full)} & \textbf{Both} \\
        \cmidrule(lr){2-4} \cmidrule(lr){5-7} \cmidrule(lr){8-8}
        & \multicolumn{2}{c}{\textbf{Vanilla}} & \textbf{Photo Rep.} & \textbf{Static} & \textbf{Static-swap} & \textbf{Dynamic} & \textbf{Mix} \\
        \multirow{3}{*}{\begin{tabular}[c]{@{}l@{}}\textbf{Verification}\\\textbf{methods $\downarrow$}\end{tabular}}
        & \multicolumn{2}{c}{120 mixed vids} & 20 vids & 110 vids & 40 vids & 90 vids & 380 vids \\
        \cmidrule(lr){2-3} \cmidrule(lr){4-4} \cmidrule(lr){5-5} \cmidrule(lr){6-6} \cmidrule(lr){7-7} \cmidrule(lr){8-8}
        & \textbf{F-score} & \textbf{Recall} & \textbf{Recall} & \textbf{Recall} & \textbf{Recall} & \textbf{Recall} & \textbf{AUC} \\
        \midrule
        \multicolumn{8}{c}{\textit{Hologram detection (no verification)}} \\
        \midrule
        \MIDVHOLO & 84 ± 5 & 82 ± 14 & 28 ± 16 & 69 ± 2 & 10 ± 5 & 2 ± 3 & 55 ± 1\\
        WSL & 95 ± 2 & 97 ± 4 & 93 ± 8 & 83 ± 6 & 18 ± 10 & 3 ± 3 & 81 ± 4\\
        \midrule
        \multicolumn{8}{c}{\textit{Generic fraud detector, leveraging any categorizable signal (e.g. print artifacts, biases)}} \\
        \midrule
        Direct Classifier & 93 ± 3 & 91 ± 5 & 72 ± 14 & 81 ± 10 & 87 ± 9 & 39 ± 17 & 93 ± 2\\
        \midrule
        \multicolumn{8}{c}{\textit{Hologram Verification (focusing more on holographic content using background subtraction)}} \\
        \midrule
        HoloVerif (ours) & 95 ± 2 & 96 ± 4 & 57 ± 19 & 93 ± 6 & 29 ± 15 & 63 ± 16 & 93 ± 2\\
        \bottomrule
        \\
        \multicolumn{8}{c}{\textbf{Ablation study}}\\
        \midrule
        \multicolumn{8}{c}{\textit{For HoloVerif (ours)}}\\
        \midrule
        Pseudo labels only & 66 ± 2 & 89 ± 12 & 84 ± 19 & 92 ± 7 & 77 ± 25 & 76 ± 12 & 61 ± 2\\
        No augmentations & 97 ± 2 & 98 ± 3 & 45 ± 19 & 94 ± 5 & 16 ± 11 & 29 ± 13 & 88 ± 3\\
        \midrule
        \multicolumn{8}{c}{\textit{ For Direct Classifier \cite{pouliquen_weakly_2024}}}\\
        \midrule
        With bg. subtraction & 98 ± 2 & 98 ± 3 & 45 ± 14 & 79 ± 10 & 16 ± 13 & 19 ± 14 & 89 ± 3\\        
        \bottomrule
    \end{tabular}
    } %
\end{table}
\clearpage

\section{Per-Attack Performance Analysis}
\label{app:per-attack}
In the main result table of the article, the proposed frauds were grouped by category. \Cref{tab:resultsdynattacks} presents recall for each individual fraud type of the proposed \MIDVDYNATTACK dataset, at 5 FPS (as in the main paper's results). Results are averages over multiple folds, with multiple seeds if possible, according to the experimental protocol described in the main paper.

\begin{table}[!h]
    \centering
    \caption{Recall performance ($\mu \pm \sigma$) at 5 FPS for each individual fraud type in the \MIDVDYNATTACK dataset. All methods were calibrated on the MIDV-Holo validation set by maximizing the F-Score as detailed in the main paper.}
    \label{tab:resultsdynattacks}
    \resizebox{\textwidth}{!}{%
\begin{tabular}{llcccc}
\toprule
& & \multicolumn{4}{c}{Recall} \\
\cmidrule{3-6}
 & Attack category & MIDV-Holo & WSL & Direct classifier & HoloVerif \\
\midrule
\multirow{7}{*}{\rotatebox{90}{Static}} & Plastic (low reflects) & 100 ± 0 & 100 ± 1 & 94 ± 10 & 93 ± 8 \\
 & Plastic (with reflects) & 91 ± 4 & 82 ± 14 & 75 ± 15 & 88 ± 10 \\
 & Paper & 83 ± 7 & 94 ± 7 & 77 ± 14 & 93 ± 5 \\
 & Laser light & 0 ± 0 & 63 ± 29 & 85 ± 33 & 100 ± 0 \\
 & LED light & 0 ± 0 & 0 ± 0 & 91 ± 16 & 100 ± 0 \\
 & Swap (2 templates) & 26 ± 10 & 19 ± 10 & 94 ± 7 & 19 ± 10 \\
 & Swap (3 templates) & 0 ± 0 & 0 ± 2 & 77 ± 23 & 55 ± 25 \\
\midrule
\multirow{5}{*}{\rotatebox{90}{Dynamic}} & Plain holo & 6 ± 6 & 2 ± 3 & 72 ± 16 & 90 ± 9 \\
 & Leaf holo & 1 ± 2 & 1 ± 2 & 37 ± 22 & 33 ± 19 \\
 & Double Sticker & 0 ± 0 & 3 ± 6 & 39 ± 34 & 72 ± 19 \\
 & Complete mask & 6 ± 13 & 6 ± 8 & 57 ± 31 & 56 ± 18 \\
 & Star and world & 0 ± 0 & 1 ± 3 & 42 ± 27 & 54 ± 22 \\
\bottomrule
\end{tabular}
} %
\end{table}
\clearpage

\section{Lightweight Implementation}
\label{app:hogsgd}

We additionally evaluated a computationally efficient variant of our method that substitutes the deep learning backbone with Histogram of Oriented Gradients (HOG) features coupled with a Stochastic Gradient Descent (SGD) classifier using log loss. This configuration maintains compatibility with the pipeline architecture, allowing for analogous image transformations while significantly reducing computational requirements. The performance metrics for this variant are presented in \Cref{tab:resultshog}.

The HOG-SGD implementation exhibited relatively consistent performance regardless of augmentation strategy. This discrepancy can be attributed to the transformation-invariant properties of HOG descriptors, which are designed to be robust against small geometric and intensity transformations. The histogram-based gradient features naturally smooth out many of the transformations applied during augmentation, effectively normalizing away differences that would otherwise benefit deep learning approaches.

\begin{table}[!h]
    \centering
    \caption{Extension of the core paper result table with 2 new simple methods evaluated at 5 FPS.}
    \label{tab:resultshog}
    \resizebox{\textwidth}{!}{%
    \begin{tabular}{lccccccc}
        \toprule
        & \multicolumn{3}{c}{\textbf{\MIDVHOLO} (test split)} & \multicolumn{3}{c}{\textbf{MIDV-DynAttack} (full)} & \textbf{Both} \\
        \cmidrule(lr){2-4} \cmidrule(lr){5-7} \cmidrule(lr){8-8}
        & \multicolumn{2}{c}{\textbf{Vanilla}} & \textbf{Photo Rep.} & \textbf{Static} & \textbf{Static-swap} & \textbf{Dynamic} & \textbf{Mix} \\
        \multirow{3}{*}{\begin{tabular}[c]{@{}l@{}}\textbf{Verification}\\\textbf{methods $\downarrow$}\end{tabular}}
        & \multicolumn{2}{c}{120 mixed vids} & 20 vids & 110 vids & 40 vids & 90 vids & 380 vids \\
        \cmidrule(lr){2-3} \cmidrule(lr){4-4} \cmidrule(lr){5-5} \cmidrule(lr){6-6} \cmidrule(lr){7-7} \cmidrule(lr){8-8}
        & \textbf{F-score} & \textbf{Recall} & \textbf{Recall} & \textbf{Recall} & \textbf{Recall} & \textbf{Recall} & \textbf{AUC} \\
        \midrule
        HOG-SGD & 80 ± 5 & 87 ± 7 & 90 ± 9 & 98 ± 3 & 37 ± 15 & 88 ± 10 & 88 ± 3 \\
        HOG-SGD (no aug.) & 82 ± 5 & 88 ± 6 & 79 ± 14 & 98 ± 3 & 38 ± 16 & 82 ± 10 & 87 ± 2 \\
        \bottomrule
    \end{tabular}
    } %
\end{table}

\end{document}